# Corpora Preparation and Stopword List Generation for Arabic data in Social Network

**Abstract**

This paper proposes a methodology to prepare corpora in Arabic language from online social network (OSN) and review site for Sentiment Analysis (SA) task. The paper also proposes a methodology for generating a stopword list from the prepared corpora. The aim of the paper is to investigate the effect of removing stopwords on the SA task. The problem is that the stopwords lists generated before were on Modern Standard Arabic (MSA) which is not the common language used in OSN. We have generated a stopword list of Egyptian dialect and a corpus-based list to be used with the OSN corpora. We compare the efficiency of text classification when using the generated lists along with previously generated lists of MSA and combining the Egyptian dialect list with the MSA list. The text classification was performed using *Naïve Bayes* and *Decision Tree* classifiers and two feature selection approaches, *unigrams* and *bigram*. The experiments show that the general lists containing the Egyptian dialects words give better performance than using lists of MSA stopwords only.

## 1. Introduction

Sentiment Analysis is the computational study of people's opinions, attitudes, and emotions towards topics covered by reviews or news [1]. SA is considered also a classification process which is the task of classifying text to represent a positive or negative sentiment [2-4]. The classification process is usually formulated as a two-class classification problem; positive and negative. Since it is a text classification problem, any existing supervised learning method can be applied, e.g., Naïve Bayes (NB) classifier.

The web has become a very important source of information recently as it becomes a read-write platform. The dramatic increase of OSN, video sharing sites, online news, online reviews sites, online forums and blogs has made the user-generated content, in the form of unstructured free text gains a considerable attention due to its importance for many businesses. The web is used by many languages' speakers. It is no longer used by English speakers only. The need of SA systems that can analyze OSN in other languages than English is compulsory.

Arabic is spoken by more than 300 million people, and is the fastest-growing language on the web (with an annual growth rate of 2,501.2% in the number of Internet users as of 2010, compared to 1,825.8% for Russian, 1,478.7% for Chinese and 301.4% for English) (http://www.internetworldstats.com/stats7.htm). Arabic is a Semitic language [5] and consists of many different regional dialects. However, these dialects are true native language forms which are used in informal daily communication and are not standardized or taught in schools [6]. Despite this fact but in reality the internet users especially on OSN sites and some of the blogs and reviews site as well, use their own dialect to express their feelings. The only formal written standard for Arabic is the MSA. It is commonly used in written media and education. There is a large degree of difference between MSA and most Arabic dialects as MSA is not actually the native language of any Arabic country [7].

There is lack of language resources of Arabic language and most of them are under development. In order to use Arabic language in SA, there are some text processing are needed like removing stopwords or Part-of-Speech (POS) tagging. There are some sources of stopword lists and POS taggers are publicly available but they work on MSA not Arabic dialect. This paper tackles the first problem of removing stopwords. Stopwords are more typical words used in many sentences and have no significant semantic relation to the context in which they exist.

In the literature, there are some research works have generated stopword lists but as far as our knowledge no one has generated a stopword list for Arabic dialects. In [8] they have proposed an algorithm for removing stopwords based on a finite state machine. They have used a previously generated stopword list on MSA. In [9] they have created a corpus-based list from newswire and query sets and a general list using the same corpus and then compare the effectiveness of these lists on the information retrieval systems. The lists are on MSA too. In [10] they have generated a stopword list of MSA from the highest frequent meaningless words appear in their corpus.

The aim of this paper is to investigate the effect of removing stopwords on SA for OSN Arabic data. Since the OSN sites and the reviews sites use the simple Egyptian dialect. The creation of a stopword list of Egyptian dialect is mandatory. The data are collected from OSN sites Facebook and Twitter [11-20] on Egyptian movies. We used an Arabic review site as well that allow users to write critics about the movies (https://www.elcinema.com). The used language by the users in the review is syntactically simple with many words of Egyptian dialects included. The data from OSN is characterized by being noisy and unstructured. Abbreviations and smiley faces are frequently used in OSN and sometimes in review site too. There is a need for many preprocessing and cleaning steps for this data to be prepared for SA. The Arabic users either write with Arabic or with franco-arab (writing Arabic words in English letters) e.g. the word "maloosh" which stands for "مالوش" which means "doesn't have". This is an Egyptian dialect word which is written in MSA as "ليس له". Sometimes they use English word in the middle of an Arabic sentence which must be translated.

We are tackling the problem of classifying reviews and OSN data about movies into two classes, positive and negative as was first presented in [2] but on Arabic language. In their work they used unigram and bigram as Feature Selection (FS) techniques. It was shown that using unigrams as features in classification gives the highest accuracy with NB. We have used the same feature selection techniques, unigrams and bigram along with NB and Decision Tree (DT) as classifiers.

We have proposed a methodology for preparing corpora from OSN which consists of many steps of cleaning, converting Franc-arab to Arabic words and translation of English words that appear in the middle of Arabic sentences to Arabic. We have also proposed a methodology of generating stopword lists from the corpora. The methodology consists of three phases which are: calculating the words' frequency of occurrence, check the validity of a word to be a stopword, and adding possible prefixes and suffixes to the words generated.

The contribution of this paper is as follows. First, we propose a methodology for preparing corpora from OSN sites in Arabic language. Second, we propose a methodology for creating a stopword list for Egyptian dialect to be suitable for OSN corpora. Third, we prepare corpus from Facebook which was not tackled in the literature for Arabic language. Fourth, tackling OSN data in Arabic language is new as it wasn't investigated much. Fourth, tackling DT classifier with these kinds of corpora is new as it wasn't investigated much in the literature. Finally, the measure of classifiers' training time and considering it in the evaluation is new in this field.

The paper is organized as follows; section 2 presents the methodology. The stopword list generation is tackled in section 3. The Experimental setup and results are presented in section 4. A discussion of the results and analysis of corpora is presented in section 5. Section 6 presents the conclusion and future work.

## 2. Methodology

The aim of our study is to prepare data from Twitter, Facebook, and a review site on the same topic in Arabic language for SA. We have chosen a hot topic on the recently shown movies in the theatres for the last festival in first of August 2014. The movies were:"الفيل الأزرق" means *"The blue elephant"*; "صنع فى مصر" means *"Made in Egypt"*; "الحرب العالمية التالتة" means *"The third world war"*; and "جوازة ميري" means *"official marriage"*. We have downloaded related tweets from twitter, comments from some movies' Facebook pages, and users' reviews from the review site *elcinema.com*.

Tweets were downloaded about the movies using the regular search of Twitter as many of the sites that retrieve tweets are closed like (searchHash, topsy). We have searched using the movies' names and downloaded all the tweets that appear at the time of search. There were many unrelated tweets downloaded as some of the movies like "صنع فى مصر" and "الحرب العالمية التالتة" can hold other meanings than the movies' title. The retrieved tweets are tweets that contain the whole words or any word either in the text or by hashtag.

The methodology we have used is very close to what was proposed in [21]. But there are some discrepancies related to the nature of the Arabic language. We have also used the removing stopwords only as a text processing technique due to lack of sources especially for Arabic dialects.

### 2.1 Corpora Preparation

The data downloaded are prepared to be able to be fed to the classifier as shown in Figure 1.

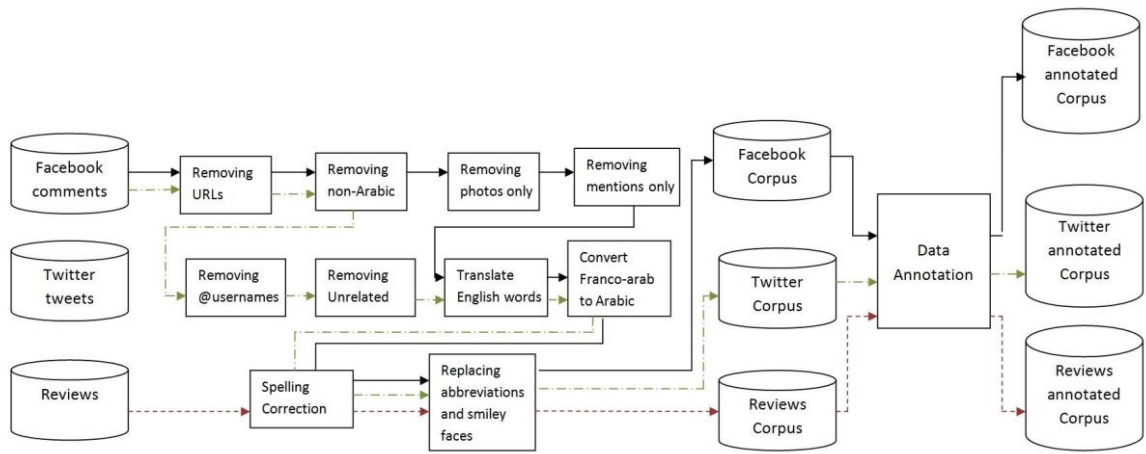

Figure 1 Arabic Corpora Preparation from Reviews, Facebook, and Twitter

The number of comments after removing the comments that contain URLs only or advertising links from Facebook was 1459. Removing comments expressed by photos only reduced them to 1415. Removing comments that contain mentions to friends with no other words reduced them to 1296. Then, after removing non-Arabic comments, they were reduced to 1261.

The final number of tweets downloaded was 1787 tweets. After removing the tweets that contain URLs only or advertising links or some who put links to watch the movie only, they were reduced to 1069. Some were links to certain scenes or related videos on Youtube. After removing unrelated tweets as the search on twitter was just by the movies' names which can imply other meanings, they were reduced to 862. Removing non-Arabic tweets reduced them to 781.

The number of reviews downloaded from the review sites was 32. The reviews needed only two steps of preparation as shown in Figure 1.

After the preprocessing, cleaning and filtering of the data, they must be annotated to be fed to the supervised classifiers. The first Experiment shows the method of annotation and the number of positive and negative data.

## 2.3 Text processing and Classification

After annotation, we have applied removing stopwords text processing technique on the three corpora with different alternatives of stopwords list which are:

 **-A general MSA list:** this list contains a combination of three published lists. The first one is a project that generated stopwords with all possible suffixes and prefixes. The other two were published in (https://code.google.com/p/stop-words/source/browse/trunk/stop-words/stop-words/stop-words-arabic.txt) and (http://www.ranks.nl/stopwords/arabic) respectively.
**-A generated corpus-based list:** this list is generated from the most frequent words from the corpora regardless of their nature.
**-A generated Egyptian-dialect list:** this list is generated from the most frequent words in the corpora that can be a stopword in addition to the Egyptian dialect stopwords that appeared in the corpora.
**-A combination of the Egyptian dialect list and the MSA list.**

Text classification is applied on the three corpora using two feature selection techniques and two classifiers as shown in Figure 2. We have used two well known supervised learning classifiers; Naïve Bayes (NB) [22] and Decision tree (DT) [23] in testing. There are many other kinds of supervised classifiers in the literature [24]. The two chosen classifiers represent two different families of classifiers. NB is one of the probabilistic classifiers which are the simplest and most commonly used classifier. DT on the other hand is a hierarchical decomposition of data space and doesn't depend on calculating probability. The test used two different feature selection (FS) techniques. These are; unigrams (BOWs) which depend on word presence; and bigrams [2].

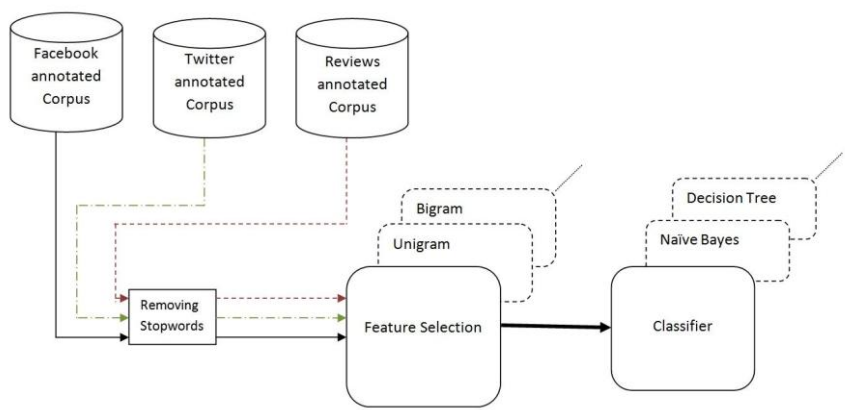

Figure 2 Text Processing and Text classification of the prepared corpora

## 3. Stopword list Generation

Stopwords are common words that generally do not contribute to the meaning of a sentence, specifically for the purposes of information retrieval and natural language processing. The common English words that don't affect the meaning of a sentence are like "a", "the", "of"…. Removing stopwords will reduce the corpus size without losing important information. In some corpora specific words could not contribute in the meaning like the word "movie" in a movie reviews corpus but means something in news corpus. This word could be considered a stopword when analyzing the movie reviews corpus.

The common strategy for determining a stopword list is to calculate the frequency of appearance of each word in the document collection then to take the most frequent words. The selected terms are often hand-filtered for their semantic content relative to the domain of the documents being indexed, and marked as a stopword list.

The English stopword list is general and contains 127 words like (all, just, being,….). In order to generate the stopword list for Arabic which is a very rich lexical language; we have done this through many steps. First, we should specify some general conditions for the word to be a stopword:

-They give no meaning if they are used alone.

-They appear frequently in the text.

-They are general words and not used specifically in a certain field.

The methodology of generating the stopword lists are shown in Figure 3. The methodology consists of three phases as illustrated in the following subsections.

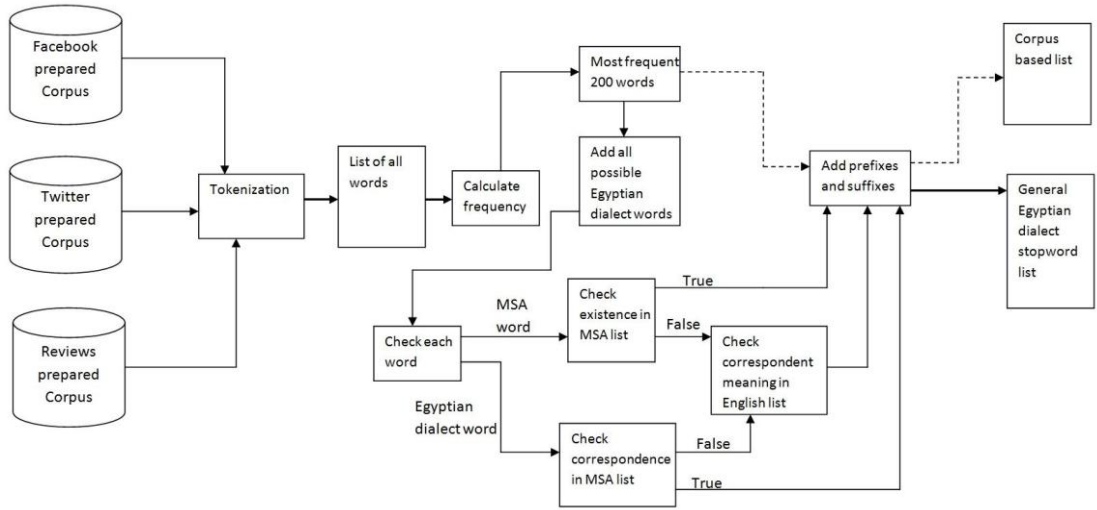

Figure 3 A methodology of generating the stopword lists

## 3.1 Calculating words frequency

The three corpora are tokenized to words. The reviews corpus give 3781 unique words, the Facebook corpus give 1451 unique words, and the Twitter corpus give 1160 unique words. This shows that despite the number of reviews are much less than the OSN corpora but they are lexically rich. After combining them together and removing the duplicates, the list of all words are 4818 words. Then we have calculated the frequency of occurrence of each word from the list of all words in the three corpora combined together.

## 3.2 The validity of words to be a stopword

To generate the corpus based list, we have taken the most frequent 200 words. These words are not all general and they are domain specific like the words "المشاهد" or "الفيلم" which means (The spectator, the movie) respectively. This list contains words in MSA and Egyptian dialect as well.

To generate a general list of Egyptian dialect stopwords, we have taken the most frequent 200 words and remove the semantically recognized words which are likely to be nouns and verbs. Then, to generate a general list of Egyptian dialect, we have added every word in the corpora in Egyptian dialect on the most frequent words that are semantically meaningless. To validate if the word is a stopword or not; if the word is a MSA words we check its existence in the MSA stopword lists. If it doesn't exist, we check its corresponding meaning in the English stopword list. If the word is in Egyptian dialect, we see its correspondence in the MSA list and if doesn't exist we check its correspondent meaning in the English stopword list. For example the word "بس", its correspondence in MSA is "فقط" and it has a corresponding meaning in the English stopword list too which is "only". On the contrary, the word "لازم" has no correspondence in the MSA list which should be "لابد" but it has a correspondent meaning in the English list which is the word "should". Therefore, it is considered a stopword. The final list of valid unique words contains 100 words.

### 3.3 Adding possible prefixes to the words

Arabic is a very rich lexical language which has a large number of prefixes and suffixes that could be added to a word to change its meaning. For example the prefix "ال" which means "the" change the word from indefinite to definite. The suffix "هم" gives the meaning of pronoun "them". We have added some frequent used prefixes to the words generated in both lists which are (ال، و، ب، ف، ل). If necessary we give pronoun suffixes which are ( ه، هم، نا، ي، ها). We have added these suffixes to possession words in Egyptian dialect like the word "بتاعى" which means (mine).

There is also some letters are written in different forms so we write any word that contains these letters' possible forms such as (ي، ى), (ه، ة), (ا، أ، إ). The last one is according to the word itself. The lists are manually revised for improper words or meaningless words.

After adding the prefixes and suffixes, the final corpus-based list contains 1061 words and can be found in (http://goo.gl/JW0jKP). The final general Egyptian dialect list contains 620 words and can be found in (http://goo.gl/263J5L).

## 4. Experimental Setup and Results

We used a HP pavilion desktop computer of model: p6714me-m. The processor is Intel(R) core (TM) i5-2300 CPU @ 2.80 GHZ; RAM is 4GB; and 64-bit operating system. We have calculated the training time using a build-in function written with python code which calculates the processing time in terms of seconds. These tests were all performed using the Natural Language Toolkit (nltk 2.0) which is implemented inside python 3.1 [25].

### 4.1 Data Annotation

The reviews from the review site were previously rated from the site. They were given a degree from 1 to 10. The ratings bigger than 5 are considered positive and less than 5 are considered negative. The ratings equal to 5 are neutral. We have annotated the reviews according to the site rating.

For the OSN data, we have manually annotated the corpora. The manual annotation was more reliable as the human being analyzing of data is better than the machine so far.. Table 1 show the number of positive, negative and neutral reviews, comments, and tweets resulted from annotation.

Table 1 Number of positive, negative and neutral reviews, comments, and tweets from Review site, Facebook and Twitter

|  | Reviews | Facebook | Twitter |
|---|---|---|---|
| No. of positive | 25 | 369 | 160 |
| No. of negative | 6 | 33 | 77 |
| No. of neutral | 1 | 859 | 544 |

### 4.2 Classifiers Preparation

We trained Naive Bayes, and Decision Tree classifiers. The classifiers were conducted with the nltk 2.0 toolkit. There are some parameters passed in to the DT classifier can be tweaked to improve accuracy or decrease training time [25].

The parameters are:

*-Entropy cutoff:* used during the tree refinement process. If the entropy of the probability distribution of label choices in the tree is greater than the entropy_cutoff, then the tree is refined further. But if the entropy is lower than the entropy_cutoff, then tree refinement is halted. Entropy is the uncertainty of the outcome. As entropy approaches 1.0, uncertainty increases and vice versa. Higher values of entropy_cutoff will decrease both accuracy and training time. It was set to '*0.8*'.

*-Depth cutoff:* used during refinement to control the depth of the tree. The final decision tree will never be deeper than the depth_cutoff. Decreasing the depth_cutoff will decrease the training time and most likely decrease the accuracy as well. It was set to '*5*'.

*-Support cutoff:* controls how many labeled feature sets are required to refine the tree. When the number of labeled feature sets is less than or equal to support_cutoff, refinement stops, at least for that section of the tree. Support_cutoff specifies the minimum number of instances that are required to make a decision about a feature. It was set to '*30*'.

### 4.3 Feature Selection

There are two Features selection (FS) techniques used in the test:

*-Unigram:* treats the documents as group of words (Bag of Words (BOWs)) which constructs a word presence feature set from all the words of an instance.

*-Bigram:* is the same as unigram but finds pair of words.

### 3.5 Results

We have made many experiments to test the effect of removing stopwords from different lists with the combination of two FS techniques and two classifiers with the three different corpora. We have made the tests on splitting 75% of the total number of the data in each corpus for training and 25% for testing data.

The standard Accuracy was used to evaluate the performance for each test. The accuracy is defined as: the ratio of number of correctly classified reviews, comment, and tweets to the total number of data.

Table 2 contains the results of the various tests we have made. The accuracy of the reviews is relatively high as the number of reviews is small and the data is highly unbalanced. The accuracy decreases when using corpus-based list on the lexically rich reviews and the general lists including Egyptian dialects give better results than the others. The timing is not changed a lot but in general it decreases when removing stopwords. The DT gives better results with Facebook data than NB as it is extremely unbalanced.

Table 2 Accuracy and training time of Sentiment Analysis on Reviews, Facebook and Twitter corpora using NB and DT classifiers with unigram and bigram as FS after removing stopwords from different lists

| Classifier | Feature selection | Removing Stopwords | Accuracy | | | Time (sec) | | |
|---|---|---|---|---|---|---|---|---|
| | | | *Reviews* | *Facebook* | *Twitter* | *Reviews* | *Facebook* | *Twitter* |
| Naïve Bayes | Unigram | *Without* | 100% | 48.04% | 78.33% | 0.050 | 0.018 | 0.015 |
| | | *Other lists* | 100% | 47.06% | 68.33% | 0.042 | 0.017 | 0.015 |
| | | *Corpus-based* | 44.44% | 53.92% | 68.33% | 0.038 | 0.014 | 0.015 |
| | | *General* | 100% | 50% | 68.33% | 0.043 | 0.016 | 0.014 |
| | | *All lists* | 100% | 48.03% | 70% | 0.041 | 0.017 | 0.014 |
| | Bigram | *Without* | 77.77% | 40.20% | 80% | 0.117 | 0.060 | 0.052 |

|  |  |  | Other lists | 88.88% | 29.41% | 68.33% | 0.090 | 0.063 | 0.053 |
|  |  |  | Corpus-based | 22.22% | 43.13% | 65% | 0.098 | 0.047 | 0.035 |
|  |  |  | General | 88.88% | 31.37% | 63.33% | 0.094 | 0.059 | 0.053 |
|  |  |  | All lists | 100% | 29.41% | 65% | 0.097 | 0.061 | 0.052 |
| Decision Tree | Unigram | | Without | 100% | 90.20% | 70% | 0.217 | 0.620 | 0.589 |
| | | | Other lists | 100% | 91.17% | 68.33% | 0.195 | 0.594 | 0.560 |
| | | | Corpus-based | 77.77% | 90.20% | 70% | 0.187 | 0.503 | 1.170 |
| | | | General | 100% | 90.20% | 68.33% | 0.196 | 0.559 | 0.530 |
| | | | All lists | 100% | 91.17% | 68.33% | 0.192 | 0.560 | 0.521 |
| | Bigram | | Without | 100% | 90.20% | 73.33% | 0.510 | 1.849 | 2.471 |
| | | | Other lists | 100% | 90.20% | 68.33% | 0.436 | 1.920 | 1.737 |
| | | | Corpus-based | 77.77% | 90.20% | 68.33% | 0.414 | 1.561 | 3.011 |
| | | | General | 100% | 90.20% | 68.33% | 0.416 | 1.787 | 0.530 |
| | | | All lists | 100% | 90.20% | 68.33% | 0.410 | 1.795 | 1.647 |

The following figures show the accuracy and logarithmic graphs of training time for each corpus. The logarithmic graphs are used to clarify the difference in timing.

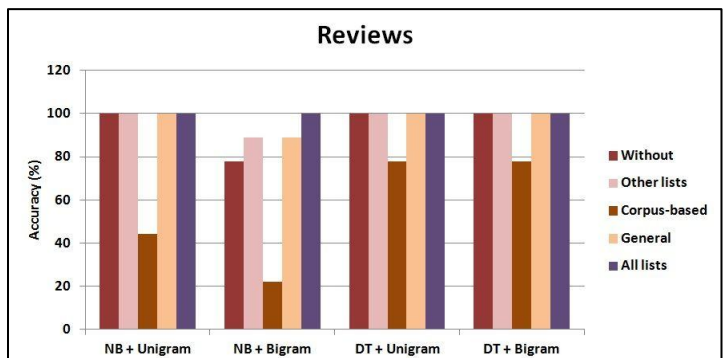

Figure 4 Classification accuracy of Reviews corpus

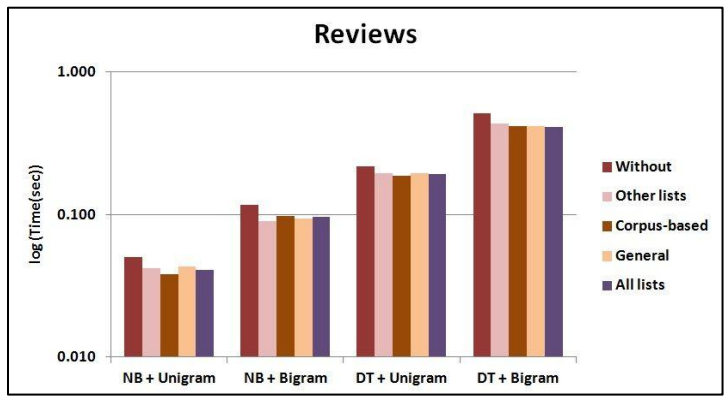

Figure 5 Classification training time of Reviews corpus

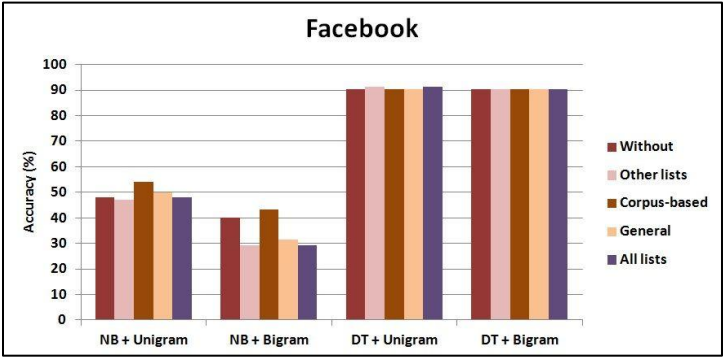

Figure 6 Classification accuracy of Facebook corpus

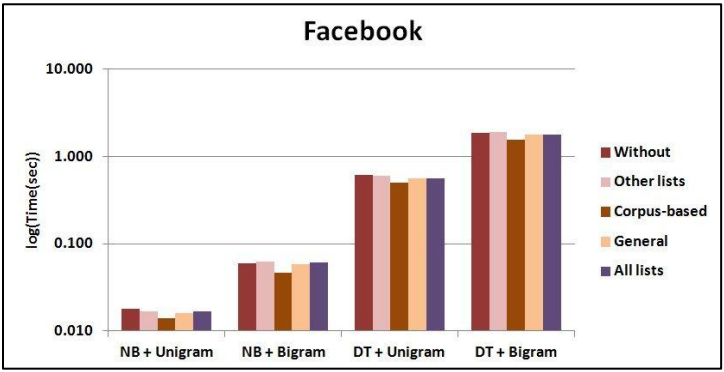

Figure 7 Classification training time of Facebook corpus

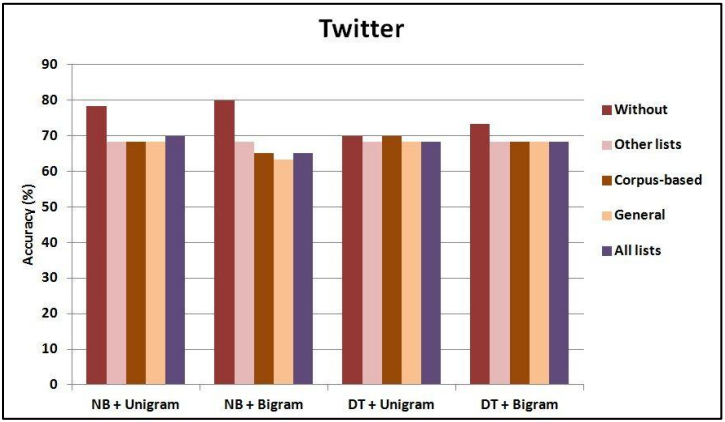

Figure 8 Classification accuracy of Twitter corpus

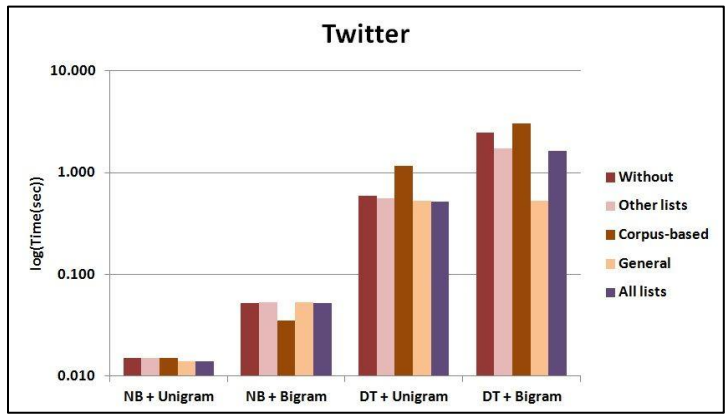

Figure 9 Classification training time of Twitter corpus

## 5. Discussion

### 5.1 Corpora Analysis

The number of neutral reviews from the review site represents 3% of the whole data. This is not a big number. We believe that people who write whole reviews on reviews sites are mainly having a complete opinion about the movie and they want to show it. They don't lean to be neutral. The number of positive reviews represents 78% of the whole data while the number of negative reviews represents 18% of the entire data. The data are obviously unbalanced since the movies were successful in this season, not many users' reviews were negative.

The number of neutral comments on Facebook represents 68% of the whole data. These are not neutral opinions on the movie. People who write in OSN are not neutral at all. The neutral comments are mainly objective sentences that don't contain any sentiments. Many were just debates between users. Some were expressing their personal feelings and some were using adjectives without specifying on whom or what. The number of positive comments represents 29% of the entire data and the number of negative comments represents 2% of the whole data which is an extremely small percentage. This is also an unbalanced data. We believe that people who access a movie page they do like it.

The number of neutral tweets represents 69% of the whole data. These are not neutral opinions on the movie too. The neutral tweets are mainly objective sentences that don't contain any sentiments. Many of the tweets were repetition of a dialogue from a movie without expressing any feelings. Others were tweets expressing the users' personal feelings like feeling excited to see the movie. The number of positive comments represents 20% of the entire data and the number of negative comments represents 9% of the whole data which is a small percentage. We believe that people who mention the movie in their tweets; do like it.

Using abbreviations and smiley faces in OSN are very frequent. There are some abbreviations were used also in Reviews. The meaning of these abbreviations and smiley faces were found from different sources on the web (Yahoo answers, Facebook emoticons sites) and translated to Arabic. For the Arabic abbreviations they were manually translated. Table 3 contains sample of Abbreviations and smiley faces found in the three corpora.

Table 3 Sample of abbreviations and smiley faces found in Facebook, Twitter, and Reviews

| Abbreviations | Facebook | Twitter | IMDB |
|---|---|---|---|
| ضحك ← هههه | Found | Found | |
| ابتسامة كبيرة ← :D | Found | Found | Found |
| قلب ← 3> | | Found | |
| مبسوط ← ^_^ | Found | Found | |

## 5.2 Specializations of Arabic Language

Words with the same meaning could be written in different correct ways like the words "هنروح، حنروح". They both give the future tense of the verb "نروح" which means "we will go". As we can notice three words in English are just written in one word in Arabic and give the same meaning. The pronouns in English are expressed in Arabic by adding a prefix letter that modify the verb especially when it is used in the middle of the sentence like "اروح، نروح" which means (I go, we go) respectively. Some prepositions and causal words are expressed in Arabic with one letter like the words "انى، لانى" which means (I am, because I am) respectively.

The many forms that the Arabic words could take are very common characteristics of MSA which make the dealing with the language is complicated. For Arabic dialects, it is a tragedy. We have a special dialect for each Arab country and different dialects in the same country. For Egyptian dialect there are many words that have no resemblance in MSA like the word "مفيش" which means (there is not). It has only a correspondent in MSA which is "لا يوجد" which are complete different words. In the OSN corpora some other dialects appear like the Moroccan word "بزاف" which means (too much) and the Syrian word "مليح" which means (good). The number of other dialects in Facebook corpus represents 1% of the whole corpus which is very small percentage. The number of other dialects in Twitter corpus represents 0.5% of the whole corpus which is extremely small percentage. There were no other dialects in reviews. They used a mix between MSA words and Egyptian dialect words as they are user reviews not formal reviews from critics.

The other phenomenon of Arab users is using the Franco-arab. This means that people use English letters for writing Arabic words like the word "de7k" which stands for "ضحك" which means (laugh). The number of Franco-arab comments in Facebook corpus represents 18% of the whole corpus which is not a big percentage. The number of Franco-arab tweets in Twitter corpus represents 3% of the whole corpus which is a small percentage. However, we have to unify the language used for the classifier to perform well. These are not even English words that have meanings so; they must be rewritten in Arabic letter. We have used the website (www.yamli.com). They give variations for each word that have to be chosen from. Sometimes the users don't even write correct words in Franco-arab. In this case the site translates the letters only which give funny Arabic words. This transformation was manually revised.

## 5.3 Results Analysis

Figure 4 shows that removing stopwords from reviews didn't change the accuracy when using general lists but decrease the accuracy when using the corpus-based list. It also shows that that unigrams are better FS than bigrams with NB. The training time decreases after removing stopwords and the training time of DT is higher than NB as shown in Figure 5.

Figure 6 shows that the accuracy of DT is much bigger than NB because the data is extremely unbalanced. There is no significant difference between unigrams and bigrams in DT but unigrams is better than bigrams with NB. Using corpus-based list increase the accuracy than using the general lists with NB but the accuracies are almost the same with DT. The training time decreases after removing stopwords and the training time of DT is higher than NB as shown in Figure 7.

Figure 8 shows that NB and DT give very close accuracies as the data is not very unbalanced. Unigrams are better than bigrams in case of NB. Using different lists didn't change the accuracy much but the general lists give good performance too. The training time decreases after removing stopwords and the training time of DT is higher than NB as shown in Figure 9.

In case of lexically rich corpus like the reviews, using the corpus-based list decrease the accuracy of classification which is similar to what [9] has found. But in case of OSN where they were not lexically rich the three lists wasn't varying the accuracies much but still the general lists containing Egyptian dialect stopwords give better results than using MSA stopwords only.

Decision Tree is a hierarchical decomposition of data space and doesn't depend on calculating probability but Naïve Bayes depends on calculating probability for the whole data. Although NB usually gives higher accuracy than DT, but this was not the case when testing these corpora. This is due to the unbalance of the data as the positive class in these cases where much bigger than the negative class. NB calculates the probability on the whole data but DT is more specifically build hierarchy decomposition of data. That is why DT is better for unbalance data as it is more specific than NB. But still DT has longer processing time than NB because it builds the hierarchical decomposition on the whole data but the difference in time is not big as the data size was not so big. In NB tests, the accuracy is better when using unigram which is similar to what [2] has found. In DT tests, unigram and bigrams give nearly similar results.

## 6. Conclusion and Future work

In this paper, we have proposed a methodology for generating a stopword list from online social network (OSN) corpora. The methodology consists of three phases: calculating the words' frequency of occurrence, check the validity of a word to be a stopword, and adding possible prefixes and suffixes to the words generated. We have generated a stopword list of Egyptian dialect and a corpus-based list to be used with the OSN corpora. We compare them with other lists. The lists used in the comparison are: previously generated lists of MSA, the corpus-based generated list, the general generated list of Egyptian dialect, and a combination of the Egyptian dialect list with the MSA list.

We have also proposed a methodology to prepare corpora in Arabic language from OSN and review site for Sentiment Analysis (SA) task. It includes the translation of English words that appear in text and the transformation of Franco-arab to Arabic words. The text classification was performed using Naïve Bayes and Decision Tree classifiers and two feature selection approaches, unigrams and bigram.

We have selected the movie reviews topic to download data about movies from three different sources (Review site, Facebook, and Twitter). The data are extremely unbalanced as the movies were successful and most of the OSN users like it and the reviewers as well. The data contain many spams like advertising URLs, debates, and using of abbreviations and smiley faces. It needed many preprocessing and cleaning steps to be prepared for classification.

Applying removing stopwords with multiple lists show that the corpus-based list negatively affects the accuracy of classification incase of reviews. Reviews are more lexically rich than OSN corpora. It also shows that the general lists containing the Egyptian dialects words give better performance than using lists of MSA stopwords only. The results of Decision tree classifier are better than Naïve Bayes classifier for these kinds of corpora. Using unigrams give better results than bigrams.

In the future we plan to try more text processing techniques on Arabic OSN data like POS tagging and try to fulfill the gap of using the Arabic dialect in the OSN data as all resources are designed for MSA. We could tackle other dialects other than Egyptian.

**References**

[1]     A. H. a. H. K. Walaa Medhat, "Sentiment analysis algorithms and applications: A survey," *Ain Shams Engineering Journal,* 2014.
[2]     B. Pang, L. Lee, and S. Vaithyanathan, "Thumbs up?: sentimentclassification using machine learning techniques," in *Proceedings of Conference on Empirical Methods in Natural Language Processing (EMNLP-2002)*, 2002.
[3]     X. Bai, Padman, R. & Airoldi, "Sentiment Extraction from Unstructured Text Using Tabu Search -Enhanced Markov Blanket," *Technical Report CMU-ISRI-04-127,* 2004.
[4]     M. Gamon, "Sentiment classification on customer feedback data: Noisy data, large feature vectors, and the role of linguistic analysis," in *Proceedings of COLING'04*, Geneva, Switzerland, 2004, pp. 841-847.
[5]     K. Versteegh, Versteegh, C., "The Arabic Language," *Columbia University Press,* 1997.
[6]     N. Habash, "Introduction to Arabic natural language processing," *Synthesis Lectures on Human Language Technologies,* vol. 3, 2010.
[7]     D. C. Mohammed Korayem, and Muhammad Abdul-Mageed2, "Subjectivity and Sentiment Analysis of Arabic: A Survey," *AMLTA 2012, CCIS 322,* pp. 128–139, 2012.
[8]     R. Al-Shalabi, Ghasan Kanaan, Jihad M. Jaam, Ahmad Hasnah, and Eyat Hilat, "Stop-word removal algorithm for Arabic language," in *1st International Conference on Information and Communication Technologies: From Theory to Applications*, Damascus, 2004, pp. 545-550.
[9]     Ibrahim Abu El-Khair, "Effects of stop words elimination for Arabic information retrieval: a comparative study," *International Journal of Computing & Information Sciences,* vol. 4, pp. 119-133, 2006.
[10]    E. M. S. A. Alajmi, R. R. Darwish, "Toward an ARABIC Stop-Words List Generation," *International Journal of Computer Applications,* vol. 46, 2012.
[11]    L. H. a. B. D. Davison, "Empirical Study of Topic Modeling in Twitter," presented at the 1st Workshop on Social Media Analytics (SOMA '10), Washington, DC, USA., 2010.
[12]    A. B. a. E. Frank, "Sentiment Knowledge Discovery in Twitter Streaming Data," 2010.
[13]    P. P. Alexander Pak, "Twitter Based System: Using Twitter for Disambiguating Sentiment Ambiguous Adjectives," in *Proceedings of the 5th International Workshop on Semantic Evaluation, ACL 2010*, Uppsala, Sweden, 2010, pp. 436–439.

[14] S. Z. Xiaohua Liu, Furu Wei, Ming Zhou "Recognizing Named Entities in Tweets," in *Proceedings of the 49th Annual Meeting of the Association for Computational Linguistics*, Portland, Oregon, 2011, pp. 359–367.

[15] M. Y. Long Jiang, Ming Zhou, Xiaohua Liu and Tiejun Zhao, "Target-dependent Twitter Sentiment Classification," in *Proceedings of the 49th Annual Meeting of the Association for Computational Linguistics*, Portland, Oregon, 2011, pp. 151–160.

[16] D. M. a. A. Funk, "Automatic detection of political opinions in Tweets," 2011.

[17] J. G. Catie Meador, "Analyzing the Relationship Between Tweets, Box-Office Performance, and Stocks," 2010.

[18] R. B. Alec Go, "Exploiting the Unique Characteristics of Tweets for Sentiment Analysis," 2010.

[19] R. B. a. L. H. Alec Go, "Twitter Sentiment Classification using Distant Supervision," 2009.

[20] P. D. Mauro Cohen, Sebastian Durandeu, Renzo Navas, Hernán Merlino And Enrique Fernández, "Sentiment Analysis in Microblogging: A Practical Implementation," presented at the CACIC 2011 - XVII CONGRESO ARGENTINO DE CIENCIAS DE LA COMPUTACIÓN, Buenos Aires, Argentina, 2011.

[21] A. H. a. H. K. Walaa Medhat, "A Framework of preparing corpora from Social Network sites for Sentiment Analysis " presented at the International Conference on Information Society (i-Society 2014), London, UK, 2014.

[22] K. N. A. McCallum, "A Comparison of Event Models for Naive Bayes Text Classification," presented at the AAAI Workshop on Learning for Text Categorization, 1998.

[23] J. R. Quinlan, "Induction of Decision Trees," *Machine Learning,* vol. 1, pp. 81–106, 1986.

[24] C. Z. Charu C. Aggarwal Ed., *Mining Text Data*. Springer New York Dordrecht Heidelberg London: © Springer Science+Business Media, LLC 2012, 2012, p.^pp. Pages.

[25] J. Perkins, *Python Text Processing with NLTK 2.0 Cookbook*. Birmingham: Packt Publishing Ltd., 2010.